\documentclass[runningheads]{llncs}
\usepackage{graphicx}
\usepackage{comment}
\usepackage{amsmath,amssymb,wrapfig} 
\usepackage{tikz}
\usepackage{color}
\usepackage{enumerate}
\newcommand{\bbR}{\mathbb{R}}
\newcommand{\bbG}{\mathbb{G}}

\newcommand{\calX}{\mathcal{X}}
\newcommand{\calU}{\mathcal{U}}
\newcommand{\calG}{\mathcal{G}}
\newcommand{\bx}{\mbox{$\mathbf{x}$}}

\newcommand{\norm}[1]{\left\lVert#1\right\rVert}

\newcommand{\calL}{\mathcal{L}}
\newcommand{\our}{\textbf{GraCIAS}}
\newcommand{\ournb}{GraCIAS}

\begin{document}
\pagestyle{headings}
\mainmatter

\title{\our: Grassmannian of Corrupted Images for Adversarial Security} 

\titlerunning{GraCIAS}
%
\author{Ankita Shukla\inst{1}\thanks{Work done as a visiting research scholar at Arizona State University, USA} \and
Pavan Turaga\inst{2}\and
Saket Anand\inst{1}}
\authorrunning{A. Shukla et al.}
%
\institute{IIIT-Delhi, India \and
Arizona State University, USA \\
\email{ankitas@iitd.ac.in, pturaga@asu.edu, anands@iiitd.ac.in}}
\maketitle

\begin{abstract}
Input transformation based defense strategies fall short in defending against strong adversarial attacks. Some successful defenses adopt approaches that either increase the randomness within the applied transformations, or make the defense computationally intensive, making it substantially more challenging for the attacker. However, it limits the applicability of such defenses as a pre-processing step, similar to computationally heavy approaches that use retraining and network modifications to achieve robustness to perturbations. In this work, we propose a defense strategy that applies random image corruptions to the input image alone, constructs a self-correlation based subspace followed by a projection operation to suppress the adversarial perturbation.  Due to its simplicity, the proposed defense is computationally efficient as compared to the state-of-the-art, and yet can withstand huge perturbations. Further, we develop proximity relationships between the projection operator of a clean image and of its adversarially perturbed version, via bounds relating geodesic distance on the Grassmannian to matrix Frobenius norms. We empirically show that our strategy is complementary to other weak defenses like JPEG compression and can be seamlessly integrated with them to create a stronger defense. We present extensive experiments on the ImageNet dataset across four different models namely InceptionV3, ResNet50, VGG16 and MobileNet models with perturbation magnitude set to $\epsilon=16$. Unlike state-of-the-art approaches, even without any retraining, the proposed strategy achieves an absolute improvement of $\sim 4.5\%$ in defense accuracy on ImageNet.

\end{abstract}

\section{Introduction}
Adversarial attacks are small imperceptible perturbations carefully crafted to modify an image that can mislead the classification ability of state-of-the-art classifiers. Robustness of deep neural networks to such attacks \cite{Fellow_ICLR2015,moosavi2016deepfool,carlini2017adv,Fellow2013} rapidly became an active area of research due to the increasing adoption rates of deep learning based systems in practical applications often with high reliability and security requirements. 
Since these applications range from autonomous driving \cite{xu2017end} to medical evaluations \cite{bar2015deep}, the robustness of these models to adversarial perturbations is a crucial aspect for their reliability.

Consequently, recent years have witnessed the development of both, adversaries \cite{moosavi2016deepfool,Fellow_ICLR2015} that challenge the robustness of deep models, as well as the design of defense strategies to mitigate their effect. White box attacks have emerged as the most challenging form of adversarial attacks, where the adversary has access to model parameters, training data as well as the defense strategy. 
In order to provide security against such attacks, various approaches like \cite{kurakin2017adversarial,Madry_ICLR2018,Fellow_ICLR2015,xie2019feature,song2018pixeldefend,Taghanaki_CVRP2019,Mustafa_ICCV2019,Fazwi_LocalLinearNIPS2019} have focused on improving the model's robustness by modifying either the network, the loss function and/or the training strategy. While these approaches have been successful to a great extent, many of them have either model or attack dependencies. Moreover, defenses that rely on adversarial training, i.e, the use of adversarially perturbed samples at train time, typically incur significant computational costs. For example, the recently proposed feature denoising approach \cite{xie2019feature} required synchronized training on \emph{128  NVIDIA V100 GPUs} for 52 hours to train a baseline ResNet-152 model on ImageNet.
While there have been efforts to reduce this training time, e.g., \cite{Fazwi_LocalLinearNIPS2019}, which achieved a 5$\times$ speed-up in training, it still required \emph{128 TPUs}. In most practical scenarios, such hardware infrastructure is not readily accessible, and even when it is, these approaches assume the access to the model and the training data. On the other hand, there exist alternate defense strategies that do not require  knowledge of the model or the training data and are more widely applicable.  

To make existing systems more robust to adversarial attacks, several approaches employ simple pre-processing or post processing strategies \cite{guo2018countering,Prakash_CVPR2018,samangouei2018defensegan,jpeg_dnn_cvpr19} that can be augmented with the deployed models directly at inference time.
For example, input transformations like JPEG compression, bit depth reduction and image quilting pre-process an image before feeding it to the network to reduce the effect of adversarial noise. The success of these transforms is due to \textit{gradient obfuscation} that results in incorrect or undefined gradients, limiting the impact of the typical gradient based white-box attacker. However, work by Athalye \emph{et al} \cite{athalye2018obfuscated} overcame this shortcoming by computing gradient approximations in such scenarios, drastically dropping the performance of many defense strategies, and often completely defeating them (0 \% classification accuracy over adversarial samples). 

 Recently, the work in \cite{BOT2019} proposed a  heuristic approach, Barrage of Random Transformation (BaRT) achieving state-of-the art defense performance on the ResNet50 model on ImageNet dataset. The defense is performed by applying a subset of image transforms in random ordering that are selected randomly from a large pool of stochastic image transformations (e.g., FFT, swirl, contrast, to name a few).
The key insight from BaRT's approach is that this \emph{compounded randomness} drastically inhibits the adversary's capability to correctly mimic the defense behavior. However, to cope up with the drastic changes in the input image due to large variations in the transforms required in BaRT, the network is fine-tuned to ensure model's familiarity with the defense at the inference time. 
On the other hand, Dubey et al. \cite{dubey2019defense} defend adversarial attacks by averaging model prediction probabilities of $k$-nearest neighbors for a given sample. Interestingly, they motivate the nearest neighbor search in an unrelated, web-scale dataset with billions of images, where the aforementioned approach of averaging the prediction probabilities is aimed at projecting the adversarial test example back on to the image manifold. These approaches either need to fine-tune the model \cite{BOT2019}, or need access to an external database \cite{dubey2019defense}, and may not strictly qualify as an input transformation based defense.

\begin{figure}[t!]
    \centering
    \includegraphics[width = 0.32\textwidth]{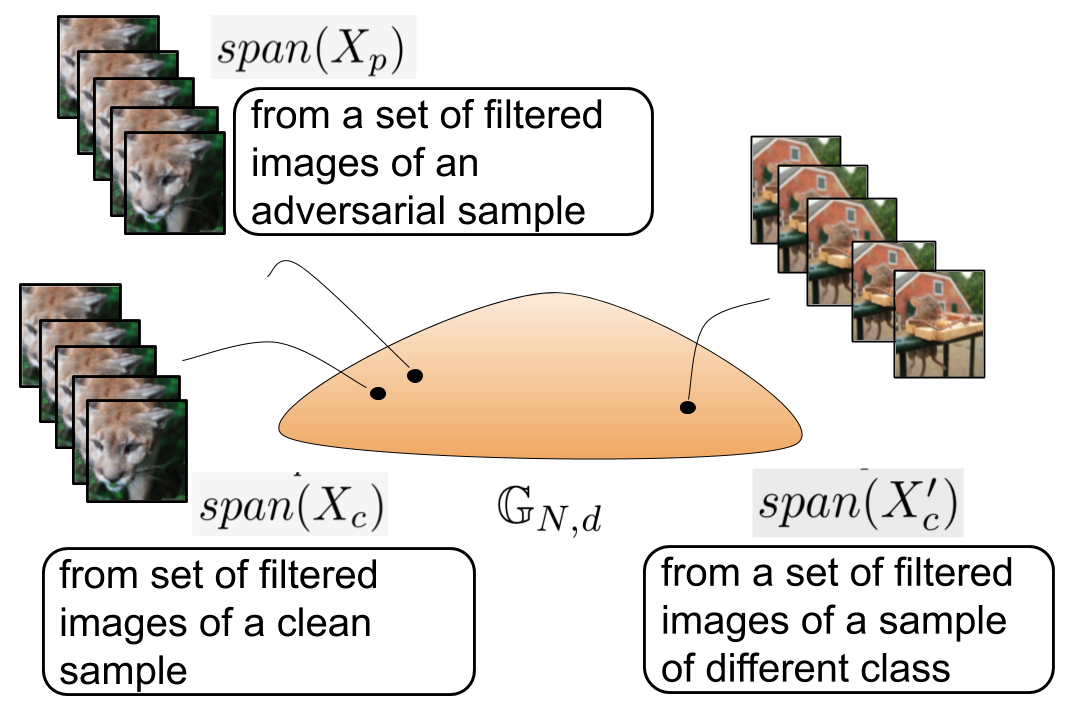}
    \includegraphics[width = 0.32\textwidth]{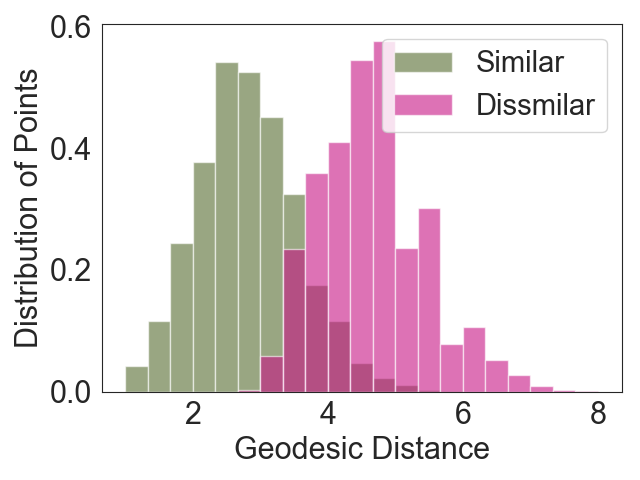}
     \includegraphics[width = 0.32\textwidth]{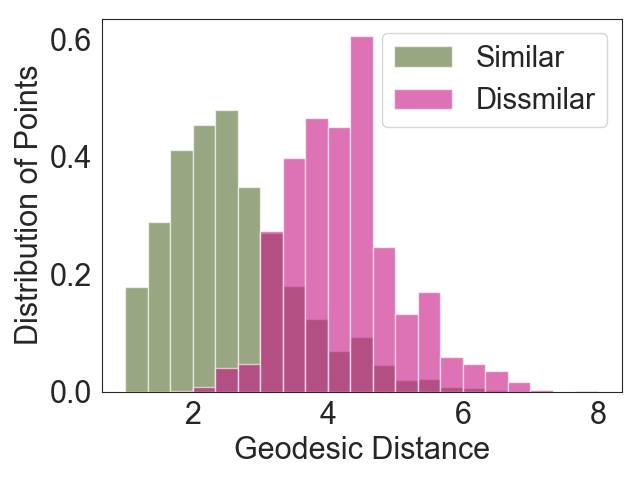}
    \caption{\textbf{Left:} Representation of subspaces as points on the Grassmannian manifold. The subspace corresponding to the perturbed sample $X_p$ lies close to the subspace of its clean sample $X_c$ counterpart. The distance between these two subspaces in shown to be upper bounded as given by Eq. (\ref{eq: final}). \textbf{Centre and Right: } The histograms show that subspaces of a pair of images of same class are closer than subspaces of image pair formed from different classes.  Given an adversarial sample, the plot highlights that the geodesic distance between clean sample subspace $\calX^l_c$ and its corresponding adversarially perturbed sample $\calX^l_p$, is such that $d(\calX^l_c,\calX^l_p) < d(\calX_c^{l'},\calX^l_p)$. Here $l$ and $l'$ represent two different classes. The plot is shown for 8000 similar ($\calX^l_c,\calX^l_p$) and 8000 dissimilar pairs ($\calX_c^{l'}, \calX^l_p$). The normalized histogram for these pairs is shown for two models on ImageNet dataset : InceptionV3 (Center) and ResNet50 (Right)}
    \label{fig:pairs_plot}
\end{figure}

To achieve our goal of devising an input transformation based defense, from the discussion above, we make two crucial observations that contribute in undermining the white-box adversary's ability to generate an attack: First is the compounded randomness in the transformations applied to the input image. Second is the access to a set of similar images (identified as neighbors in the web-scaled database in case of Dubey \emph{et al} \cite{dubey2019defense}) to have an averaging or smoothening effect over predictions, leading to more accurate predictions for an adversarial sample. Contrary to \cite{BOT2019,dubey2019defense}, we take an approach that simply relies only on a given sample and leverages benefits from both, the compounded random transformations as well as smoothing. However, the random transforms and smoothing are performed in a \emph{principled manner} that reduces the impact of adversarial noise, without significant changes in  the image. This makes our approach a model-agnostic, inference-time defense, that does not rely on additional data or training to achieve the desired goal of adversarial security.  

Our proposed approach, \textbf{Gra}ssmannian of \textbf{C}orrupted \textbf{I}mages for \textbf{A}dversarial \textbf{S}ecurity (GraCIAS) applies a random number of randomized filtering operations to the input test image. These filtered images provide a basis for a lower dimensional subspace, which is then used for smoothing the input image.
Due to a principled structure of generated image corruptions used for defining the subspace, it permits projection and reconstruction of the input image without substantial loss of information. Furthermore, we can interpret these subspaces as points on the Grassmann manifold, which
permits us to derive an upper bound on the geodesic distance between the subspaces obtained by filtering a clean sample and its adversarially perturbed counterpart. This is also supported with empirical analysis, which suggests that the geodesic distances between the subspaces corresponding to clean and adversarial examples belonging to the same class are smaller than those corresponding to examples from different classes. Figure \ref{fig:pairs_plot} illustrates the subspace representation and shows that the distribution of geodesic distances computed between subspace pairs of the same class and that of different classes are reasonably separable. 
This observation is central to our approach and validates that our choice of filters ensure a low-dimensional subspace that is representative of the test sample's original class, and serves as an appropriate smoothing operator for the input samples. Through extensive experiments on the ImageNet dataset, we show the effectiveness of GraCIAS on several models under attack of various strengths. We summarize our contributions below: 

\begin{itemize}
    \item The proposed input transformation based defense achieves state-of-the-art results on ImageNet dataset for ResNet50, InceptionV3, VGG16 and MobileNet  models under different attacker strength in white box attack scenario.
    \item As opposed to state of the art randomized input transformation approaches, $\our$  benefits not only from its intrinsic random parametrization, but also from the theoretical motivation that suggests retention of task-relevant information and suppression of adversarial noise. 
    \item Due to its simplicity and computational efficiency, the proposed defense can be integrated with existing weak defenses like JPEG compression to create stronger defenses, as shown in our experiments.
\end{itemize}

\section{Previous Work}
The vulnerability of neural networks to adversarial perturbations has led the growth of a large number of defense strategies. 
Given the large volume of work in this area, we categorize the literature into two broad groups and briefly review recent developments in them.

\noindent\textbf{Robust Training and Network Modification.} Robust training refers to strategies that retrain a model with an augmented training set. The most popular strategies use adversarial training \cite{Madry_ICLR2018,Fellow_ICLR2015,kurakin2017adversarial}, where adversarially perturbed samples are included in the training set on the fly, i.e., during the model training. These approaches, while effective, are very computationally expensive, as the attacks have to be regenerated multiple times during the entire training process. On the other hand, approaches \cite{xie2019feature,Taghanaki_CVRP2019} modify network architecture to achieve adversarial robustness. Xie \emph{et al} \cite{xie2019feature} added feature denoising blocks in the model to circumvent the impact of noisy feature maps caused due to adversarial perturbations at the input. Whereas, \cite{Taghanaki_CVRP2019} transformed layerwise convolutional feature map into a new manifold using non-linear radial basis functions. However, both robust training and network modification, not only require access to model parameters and training data, but also necessary computational resources to perform the retraining, which may be nontrivial to obtain. This requirement poses a bottleneck in securing the already deployed systems for various applications that are built on deep learning models. Therefore, in case of restrictions on computational resources or access to model parameters, add-on defenses in the form of pre-processing blocks at the input or output of a network are viable.

\noindent\textbf{Input Transformations.} The limitations of the previous category are addressed by the input transformation based approaches, which aim to denoise the image before feeding it to the network for classification. Most transformation based defenses like the ones proposed in \cite{guo2018countering}, e.g., image compression, bit depth reduction, image quilting etc., lead to \emph{obfuscated gradient}, a way of gradient masking that gives a false sense of security. Such defenses have limited robustness on a given model under powerful attacks \cite{athalye2018obfuscated}. Other approaches that have been successful to some extent under strong attacks like the Barrage of Random Transforms (BaRT) \cite{BOT2019} follow a highly randomized approach to choose at random from an \emph{enormous} pool of transformations, making it difficult for the attacker to break. BaRT requires the model to be finetuned on the input transformations to reduce the drop in performance on clean samples. Another defense in this category \cite{jpeg_dnn_cvpr19} improves the standard JPEG compression to tackle the adversarial attack without significant drop in performance of clean samples. Other approaches exist that assume access to the training set. They either learn valid range spaces of clean samples, or use the training set to approximate the image manifold, on which the input image is projected  \cite{samangouei2018defensegan,song2018pixeldefend}.  

Our proposed approach \textit{\ournb} is also an input transformation based approach. We emphasize that unlike most existing approaches, our defense is agnostic to model and the training set used.



\section{Proposed Approach}
\begin{figure*}[h]
    \centering
    \includegraphics[width=\textwidth]{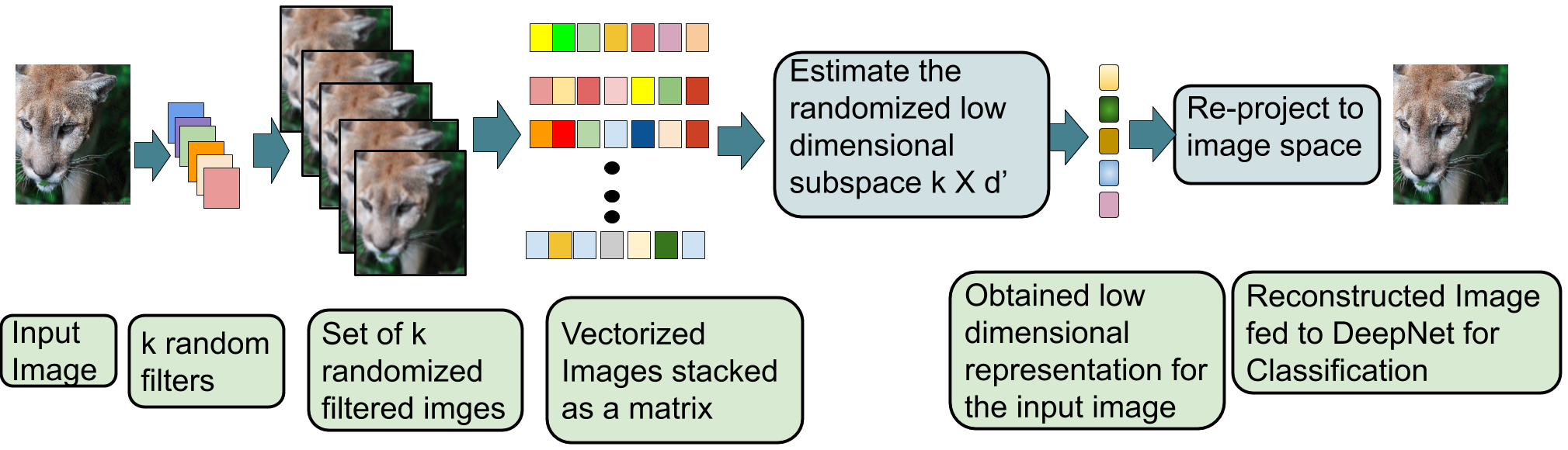}
    \caption{An overview of $\our$ defense applied on an adversarial sample. The number of $k$ random filters are used for creating a set of corrupted images. These image are used to estimate a random $d$ dimensional subspace that is used for obtaining the low dimensional representation followed by re-projection to image space to obtain a rectified image. }
    \label{fig:app}
\end{figure*}
Given a trained deep network, an adversary can add an imperceptible perturbation to the input sample that forces the model to make a wrong prediction. For a given sample $\bx$, an adversary generates a sample $\hat{\bx} =\bx +\delta$ such that its label $l(\cdot)$ does not match that of the original sample, i.e., $l(\bx) \neq l(\hat{\bx})$. Thus, the objective of an attacker can be understood as follows
$\arg \max \calL(\bx+\delta,y) \;\;   s.t. \norm{\delta}_p< \epsilon $

Here, $y=l(\bx)$ is the ground truth label of sample $\bx$, $\delta$ is the added perturbation and $\epsilon$ is the perturbation bound.
\\
\noindent \textbf{Design Goal:}
The goal of an input transformation based defense strategy is to `clean' an adversarial sample before feeding it into a classification network. The `cleaning' should reduce mis-classifications due to the perturbation, while maintaining performance on clean samples. In this work, we propose an input transformation-based \emph{inference time} approach that is \emph{simple} and \emph{methodically randomized} to achieve effective defense.

We strive to achieve this design goal and develop our defense strategy \textit{GraCIAS} combines simple randomized corruptions, subspace projections and a geometric perspective on the input transformations. 
Figure \ref{fig:app} shows an overview of our approach, which we describe in detail in the following sections.
\subsection{Proposed Defense Strategy}\label{sec:proposed_ds}
The process of generating the transform to rectify an adversarial sample $\bx_p$ is described as follows.

\noindent\textbf{Image Set for Subspace Approximation.} The aim of an input defense strategy is to find a transform that can estimate a clean sample from a given adversarial sample. As opposed to Barrage of random transforms \cite{BOT2019}, we focus on developing a random transform that is minimal without compromising the effectiveness of the transform. Our transform comprises of a projection step from image space to a low dimensional space and a reconstruction step to project back to image space. The first step is to generate a database required for estimating the subspace for projecting the adversarial sample. To this end, we use random  image filtering to generate several noisy versions of a given adversarial sample. For $k$ such random filters, the set of $k$ blurred images can be written as
\begin{align} \label{eq:ptrb_images}
  X_p =\{\bx_p * h_1, \bx_p*h_2,....\bx_p*h_k \} 
\end{align}
This step essentially corresponds to mixing a uniformly structured noise to the non uniform noise in the image caused due to adversarial perturbation. This mixing is achieved by multiple convolutions with kernels $\{h_1, h_2 ...h_k\}$ that have uniformly distributed weights normalized to have unit $\ell_1$-norm \emph{i.e.} $h_i(m,n)$ $\sim$  $\calU[0,1]$ and $||Vec(h_i)||_1 =1$, where $Vec()$ is the vectorization operator and $m,n$ are the indices of the kernel.
In addition to the random kernels, the \emph{number} of such kernels, $k$ to create the set of corresponding $k$ images is  also picked at random from a fixed range of values $k_{range}$. This choice of randomizing filter kernels and their number is driven by our goal of increasing randomness in the defense.
Such random filters are more effective than other filters like gaussian blur, where their parametric nature is much easier for attacker to approximate. 
\\
\noindent \textbf{Randomized Subspace for Projection.} 
As $X_p$ is derived from random filtering of the input image itself, the span of its elements is likely to retain some information relevant for the end task. So we simply find an orthogonal basis for the subspace spanned by $X_p$. 
The projection of the input image on to this subspace has a blurring effect on the adversarial perturbation mixed with random noise. In the absence of the adversarial perturbations, i.e., for clean samples, a similar blurring is expected along with retention of task-relevant information. Computing the basis for the subspace is computationally inexpensive even for high resolution datasets like ImageNet as the set of corrupted images is fairly small.

\noindent \textbf{Re-projection into Image Space.} The final step is to reconstruct the image from the low dimensional mapping obtained in the previous step. As in PCA based reconstruction using the low dimensional mapping and the inverse transform, we obtain the restored image. The basis of the subspace will capture relevant image content in the first few leading principal components while the noisy components are captured by the later. Thus, restoring to a low dimensional subspace will filter out the noisy components. However, using a fixed dimension for subspace can be easily estimated by a white box adversary, weakening the defense effectiveness. Therefore, the dimension of the subspace is defined based on retaining a specific value of variance in the data. The variance value is selected randomly from a predefined range of values. This adds another level of randomness in choosing the subspace dimension, making the defense effective in the the presence of an adaptive attacker that is aware of the defense strategy.

\subsection{Validity of Proposed Subspace}
We now present our analysis to show that the subspace estimated with an adversarial example is close to subspace created with clean image counterpart. Our theoretical result is based on bounds obtained on the geodesic distance between the subspaces constructed by using the clean sample $\bx_c$ and the adversarially perturbed sample $\bx_p$.

The column space of $X_p$ and $X_c$ are the subspaces containing the set of convolutions of $\bx_p$ and $\bx_c$ with random kernels $h_i$ i.e.,  $span(X_p)$ and $span(X_c)$ respectively. We represent the subspaces as $\calX_p$ and $\calX_c$ for perturbed and clean image respectively, which are both $d$-dimensional subspaces in $\bbR^{N}$, where $N=3mn$. 
Now, we want to compare these linear $d$-dimensional subspaces in $\bbR^{N}$ for which we make use of the Grassmann Manifold, $\bbG_{N,d} $, which is an analytic manifold, where each point represents a $d$-dimensional subspace in $\bbR^N$ regardless of the specific bases of the subspace. 
The distance between the two subspaces is then given by the geodesic distance between the points on the Grassmann manifold. The normalized shortest geodesic distance is defined as follows
\begin{align}
   d_{ng}(\calX_c, \calX_p) = \frac{1}{D}d_{g}(\calX_c,\calX_p) 
\end{align}
Here, $D$ is the maximum possible distance on $\calG_{N,d}$ \cite{ji1987,li2014grassmann}.
It was shown in \cite{ji1987}, that this normalized distance is upper bounded by the following expression
\begin{align}\label{eq: original}
   d_{ng}(\calX_c, \calX_p) & \leq \norm{X_c}_F \norm{X_c^{\dagger}}_2 \frac{\norm{\Delta X}_F}{\norm{X_c}_F}
  \\ & \leq \norm{ X_c^\dagger}_{2} \norm {\Delta X}_F
\end{align}

Here, $\norm{X_c^\dagger}_{2}$ is the spectral norm of pseudo inverse of $X_c$, $\Delta X= X_c - X_p$ and $\norm{\cdot}_{F}$ denotes the Frobenius norm. Rewriting, \ref{eq: original} with squaring on both sides as 
\begin{align}\label{eq: sq_org}
   d_{ng}^{2}(\calX_c, \calX_p) \leq \norm{X_c^\dagger}_{2}^2 \norm{\Delta X}_{F}^{2} 
\end{align}
Here, $ \norm{X_c^\dagger}_{2}$ corresponds to the smallest eigenvalue of the $X_c^\dagger$ \emph{i.e.} inverse of the smallest singular value of $X(\bx_c)$. Hence, we can write the following
\begin{align}\label{eq:eigen}
    \norm{X^\dagger}_2 = \frac{1}{\sigma_{min}(X)}
\end{align}
In case of natural images the $\sigma_{min}$ is non-zero. The eigenvalues of a blurred image decay faster than the clean images, as the blurred images are dominated by low frequency components \cite{hansen2006deblurring}. Similarly, the other factor in the right side of  \eqref{eq: sq_org} is given by 
\begin{align} \label{eq:bttb}
    \norm{\Delta X}_F^2 = \sum_i \norm{H_i}^2 \norm{\delta}^2
\end{align}
Here, $H_i$'s are BTTB matrices that are full rank under zero boundary condition for convolution and $\delta$ is the adversarial noise added to the clean image. Thus, substituting \eqref{eq:eigen} and \eqref{eq:bttb} in \eqref{eq: sq_org}, we get
\begin{align}\label{eq: final}
   d_{ng}^{2}(\calX_c, \calX_p) \leq \frac{1}{\sigma_{min}(X)} \sum_i \norm{H_i}^2 \norm{\delta}^2
\end{align}

The bound in the above equation establishes that the subspaces of clean and adversarial sample are in close proximity, as long as the singular value term is bounded above. To show the latter, we adopt the following approach. It is not possible to provide a general bound on $\sigma_{min}(X)$ without assuming something about natural image statistics. On the other hand, it is easy to see that $\sigma_{min}(X)$ is small only for pathological examples. Consider the case of $\sigma_{min}(X) = 0$. This happens if and only if the columns of $X$ are linearly dependent. That is, there must exist non-zero scalars $\alpha_i$ such that:
\begin{align}
    x_c = \sum_i \alpha_i \left(\bx_c * h_i\right)
\end{align}

Using simple Fourier transform arguments, it can be shown that the above happens only under pathological cases such as when $\bx_c$ is a constant-image, or the filters $h_i$ are all just plain delta functions. For any general situation, $\sigma_{min}(X)$ is finite, although a general lower bound is hard to find. 


The random filtering and the subspace projection reduces the effect of adversarial noise. Further, in the above discussion, we have established that the subspaces derived from clean samples and those from their adversarially perturbed counterparts are close to each other. If such a subspace $\calX_c$ is representative of the clean sample, i.e., if it captures information relevant to the end task, then a nearby subspace like $\calX_p$ is also likely to retain similar information. Therefore, a projection and reconstruction operation on such a subspace ($\calX_c$ or $\calX_p$) will achieve our objective of reducing adversarial noise and yet retaining relevant information. In addition to this proximity guarantee in (\ref{eq: final}), our empirical analysis of geodesics in Fig. \ref{fig:pairs_plot} shows that the geodesic distances between subspaces obtained from an adversarial sample is closer to its clean counterpart than that of another clean image from a different class. 

This observation along with our proximity result is the basis of our main hypothesis: The subspaces resulting from our randomly corrupted versions of the input image are sufficiently representative to retain task-relevant information and yet are effective transformations in attenuating adversarial noise. In the following sections, we validate this hypothesis through extensive empirical evaluation and analysis experiments. 



\section{Results}
In this section, we firstly evaluate our defense strategy as pre-processing at inference time on adversarial samples generated with different perturbation magnitude as well as attacker's knowledge. Secondly, we also present ablation experiments to thoroughly evaluate the choice of parameters used in the proposed defense strategy. Now, we list down the dataset, models and attack methods used to evaluate the performance of our defense strategy.

\subsection{Experimental Setup}
\noindent\textbf{Datasets and Models.} We present a series of experiments to evaluate the effectiveness of our defense strategy.


\noindent \textbf{ImageNet-50K}\cite{ImageNet} dataset contains images of different sizes distributed along 1000 categories. The images are processed to $256 \times 256$ dimensions encoded with 24 bit color. The validation set consists of 50,000 images. We represent this entire set as ImageNet-50K and a subset of first 10,000 images as \textbf{ImageNet-10K} for our experiments.  For ImageNet \cite{ImageNet}, we evaluate the performance on InceptionV3 \cite{Inception16}, Resent50\cite{ResNet}, MobileNet \cite{MobileNet} and VGG16 \cite{simonyan_VGG2015}. We used pre-trained models available in Tensorflow. 

\noindent\textbf{Comparison with other Approaches.} We compare our defense strategy with several input transformations that are summarized below:

\textbf{JPEG compression, BitDepth reduction \cite{guo2018countering} and JPEGDNN \cite{liu2019feature}}: These defenses are applied to an image only at the inference time. Similar to our approach, these approaches are model agnostic and hence can easily be integrated with existing systems. We used the framework of  \cite{athalye2018obfuscated} and the authors' implementation from github for evaluating these defenses under different attack scenarios. The JPEG defense is performed at a compression quality level of 75 (out of 100) and the images are reduced to 3 bits for BitDepth defense in all our experiments. For JPEGDNN, we used default parameters available with the authors' implementation that is available on github.

 \textbf{Barrage of Random Transformation \cite{BOT2019}} We compared state-of-the art performance of BaRT on ResNet50 model with our defense strategy on ImageNet50K. As the authors' implementation is not publicly available, we use results as reported in their paper in our comparisons. While we outperform BaRT on accuracy of attacked images by a significant margin, our performance on clean images is lower. However, it is important to point out that BaRT's model is fine-tuned for an addition 100 epochs with their transformed images, whereas we use the original model.

\noindent\textbf{Parameters in GraCIAS.} In our defense strategy there are two steps that require random parameter choice. First is the \emph{number} of filters $k$ to create the set of corrupted images as given by \ref{eq:ptrb_images}. To limit the computational cost, it is chosen from a fixed range $k_{range}$. This range is set to $\calU[10,60]$ i.e. a minimum of 10 corrupted images to a maximum of 60.
The other random parameter that adds to the robustness of the proposed defense strategy is the dimensionality of the subspace defined on the set of the set of images mentioned earlier. The dimension is computed based on the $\%$ of data variance retained while computing the PCA basis. In order to avoid information loss and drastic image changes,  the variance is selected between 60 \% to 95\% at random. The third parameter is the kernel size for the filtering operation. The filter size is fixed to $7 \times 7$ for ImageNet dataset.
\subsection{White Box Attacker}
 The white box adversary has access to model parameters, training data and trained weights to generate adversarial samples.  We evaluate the performance of different models under FGSM \cite{Fellow_ICLR2015} and PGD \cite{Madry_ICLR2018} attacks with $L_{\infty}$ distance. The different iterations of PGD attack are denoted by PGDk. For example, PGD with 10 iterations is denoted with PGD10. 
 
We evaluated the performance of InceptionV3 model under FGSM and PGD attack. The results are present in Table \ref{tab:all_model_fgsm_pgd_imagenet} with perturbation value of  $\epsilon=16$.  The table also presents results corresponding to standard JPEG compression and Bitdepth reduction that have been shown to be effective when defense is not known to the attacker. These transformations are applied at the inference time to transform the sample before feeding it to a pre-trained InceptionV3 network. The recent work on DNN guided JPEG compression \cite{liu2019feature}
was shown to achieve state-of-the-art results when compared with other input transformations like image quilting, standard JPEG and bitdepth. While JPEGDNN performs well in low perturbation setting, the performance drops for high perturbation value of $\epsilon =16$ as reported in the table. Our approach outperforms previously best known result using JPEGDNN as input transformation.
\begin{table}[t]
    \centering
     \caption{ImageNet 10K validation set: Comparison of different input transformation based defense on InceptionV3 model. The table reports defense classification accuracy under FGSM, PGD40 and PGD100 attacks with an attack magnitude of $\epsilon =16$}
    \begin{tabular}{|c|p{1.9cm}p{1.9cm}p{1.9cm}p{1.9cm}p{1.9cm}|}
        \hline
        Attack & \multicolumn{5}{c|}{ImageNet10K (InceptionV3)} \\ \cline{2-6}
        & No def & JPEG & BitDepth &JPEGDNN & GraCIAS \\ 
        &  &(ICLR'18) &(ICLR'18) &(CVPR'19) &(Our) \\ \hline
         FGSM & 22.88 & 25.32&24.76  &26.21 & \textbf{37.63}\\
         PGD40 & 0.0 &0.65 &0.27& 3.52 & \textbf{42.49}   \\
         PGD100 &0 &0.0 & 0.23&3.01&\textbf{44.32}  \\
          \hline
    \end{tabular}
    \label{tab:all_model_fgsm_pgd_imagenet}
\end{table}{}
\vspace{-0.5cm}
 \begin{figure}
     \centering
     \includegraphics[width=0.43\textwidth]{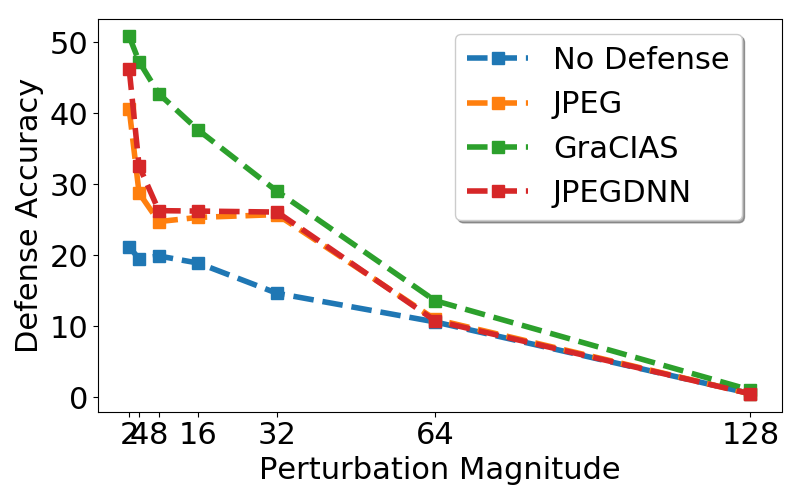}
     \includegraphics[width =0.43\textwidth]{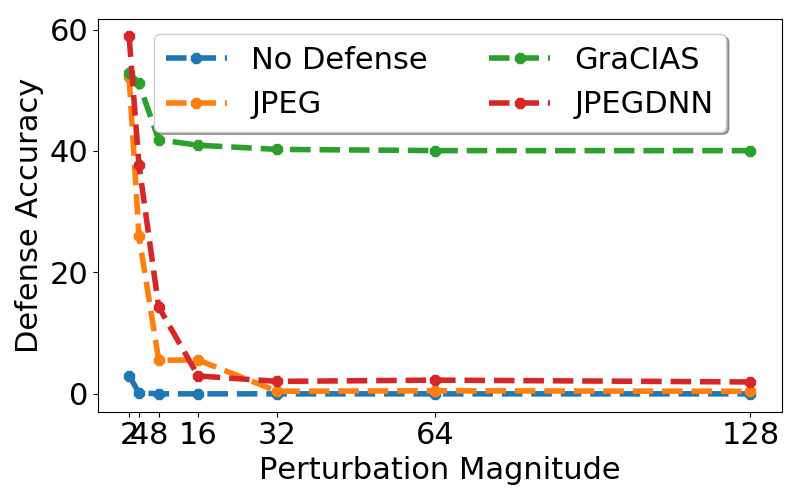}

     \caption{Performance Comparison of various defenses across different magnitude of $\epsilon$ using (Left) FGSM and (Right) PGD10 attacks on InceptionV3 model.}
     \label{fig:fgsm_pgd}
 \end{figure}
 \vspace{-0.3cm}
 \begin{wrapfigure}{r}{0.6\textwidth}
    \centering
    \includegraphics[width = 0.6\textwidth]{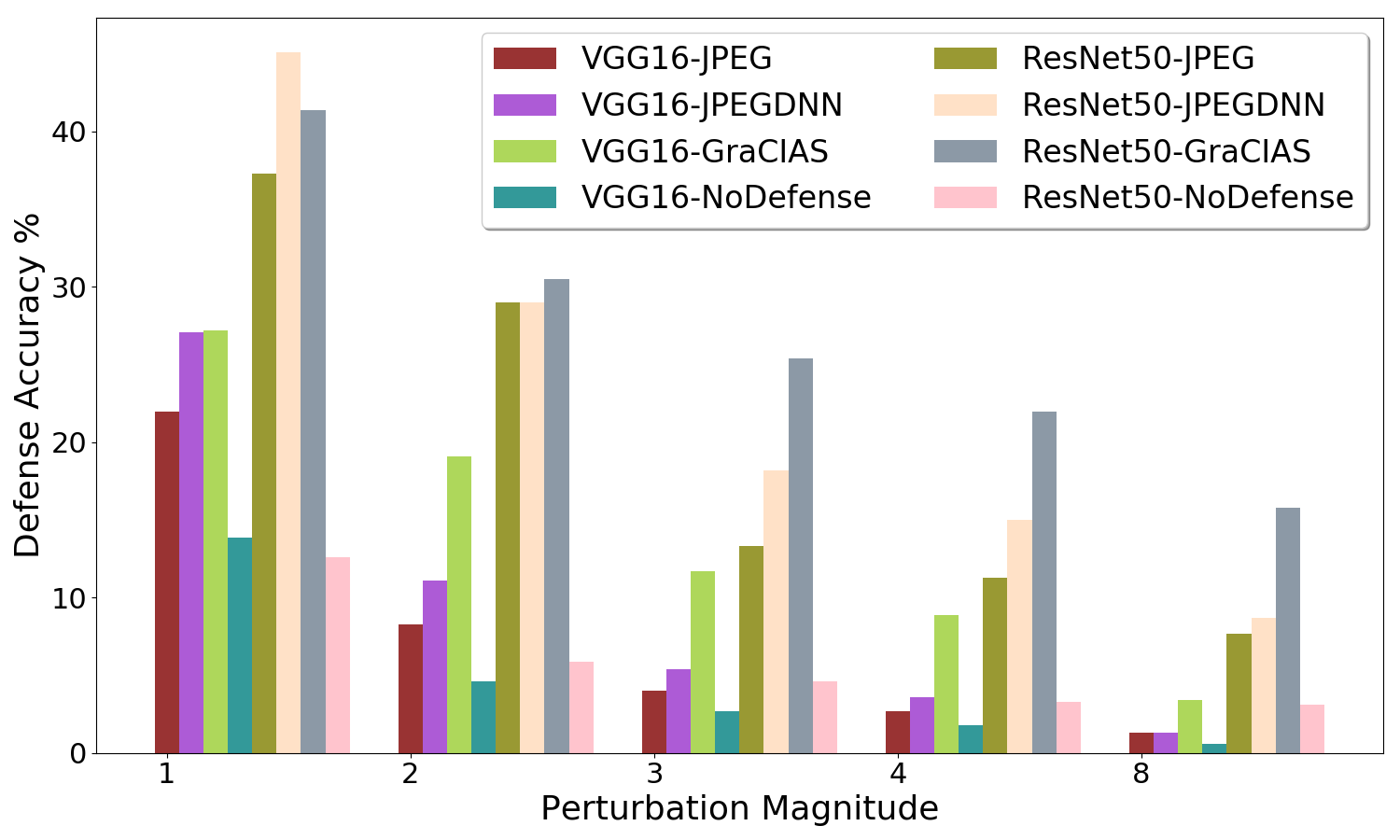}
    \caption{Performance comparison of different defense strategies under different magnitude of FGSM attack on ImageNet dataset for VGG16 and ResNet50 models. }
    \label{fig:diff_fgsm_vgg_resent}
\end{wrapfigure}{}

 \noindent \textbf{Performance across different Perturbation Magnitude.}
We evaluated the performance of our defense strategy over a range of perturbation magnitude for both PGD and FGSM attacks and compared with state-of-the-art approaches. The results are shown in Figure \ref{fig:fgsm_pgd} for InceptionV3 and for VGG16 and ResNet50 in Figure \ref{fig:diff_fgsm_vgg_resent}. The results indicate that \ournb~ sustains for larger range of perturbation before dropping at large magnitudes of perturbations.
\subsection{Adaptive Adversary}
\begin{figure}[h!]
    \centering
    \includegraphics[width =0.46\textwidth]{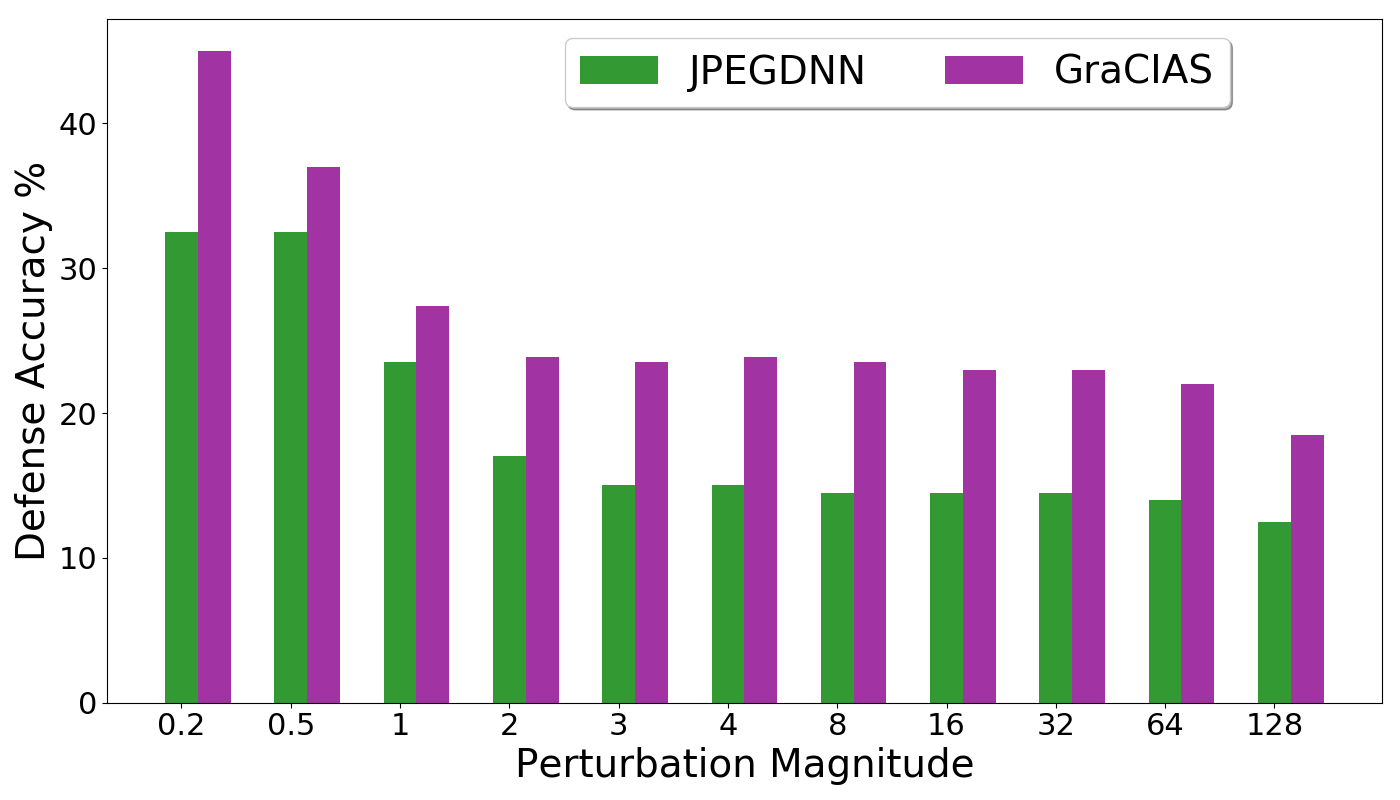}
     \includegraphics[width=0.46\textwidth]{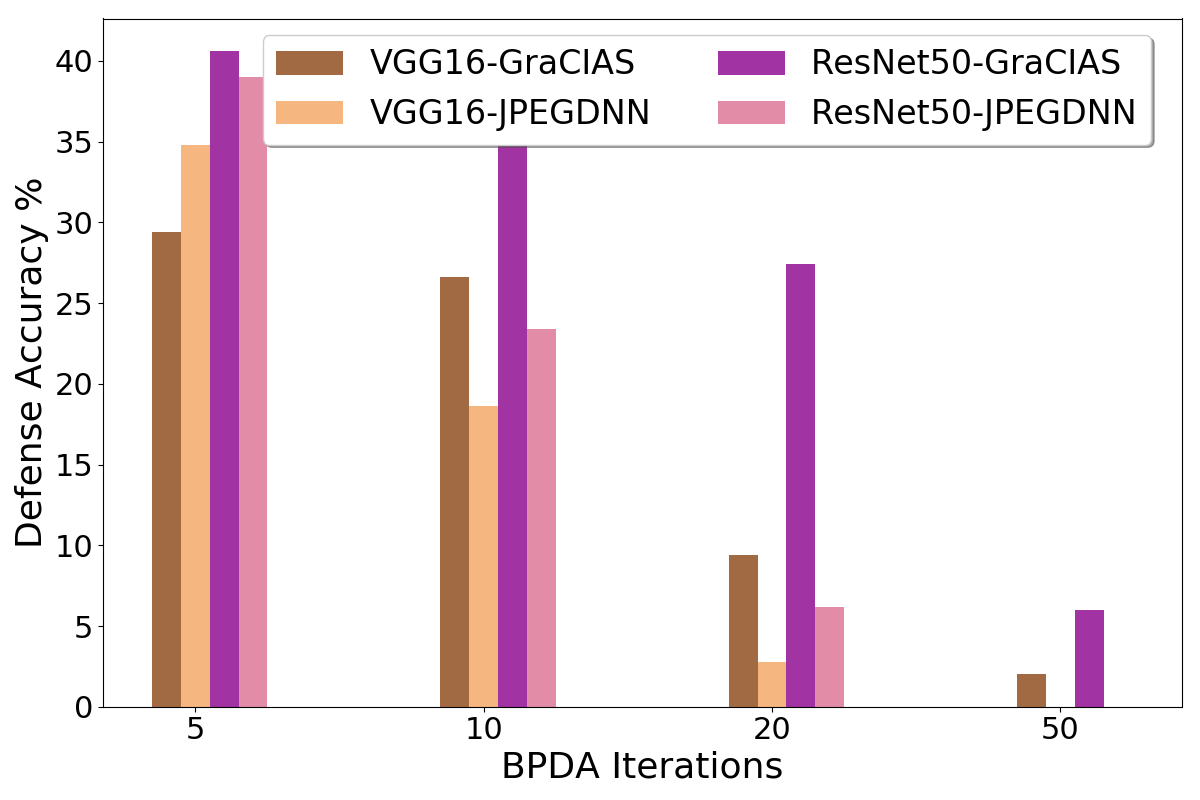}
    \caption{\textbf{(Left)} InceptionV3 model under BPDA attack with different perturbation magnitude on ImageNet dataset. The plot highlights that GraCIAS achieves state of the art results over previously reported with JPEGDNN. \textbf{(Right)} Performance of defense accuracy on ResNet50 model under different iterations of BPDA attack with $\epsilon =8$. While both ResNet50 and VGG16 are completely defeated at increased attacker's strength, GraCIAS still achieves non-trival defense performance.}
    \label{fig:bpda_eps}
\end{figure}

The recent work \cite{athalye2018obfuscated} showed that existing input transformation based defense strategies can be attacked with a strong attacker that has access to the defense strategy as well. A non-differentiable defense strategy can be defeated with BPDA (Backward Pass Differentiable Approximation) that approximates the gradient with identity while backpropagating through the transformation layer to develop strong attacks. We show that the proposed defense strategy can withstand such attacks as opposed to existing defense strategies.  The results in Table \ref{tab:imagenet_bpda} indicate that the proposed approach achieves state-of-the-art results on both InceptionV3 and ResNet50 models with approximately  $4 \%$ improved over state-of-the-art results. We also validated the efficacy of our approach across different perturbation magnitude as well as attack iterations in Fig \ref{fig:bpda_eps} and show that \ournb~ achieves non-trival accuracy as opposed to current state of the art input defense JPEG DNN.
 \begin{table}[h]
    \centering
    \caption{ImageNet Validation Set: Performance Comparison of defense classification accuracy under BPDA attack ($\epsilon = 16$, iteration 40) on InceptionV3, ResNet50, MobileNet and VGG16 models. \textbf{*} indicates that the results are quoted from the respective paper, in the absence of open source implementation.}
    \begin{tabular}{|c|p{1.5cm}|p{1.8cm}p{1.8cm}p{1.8cm}p{1.8cm}|}
    \hline
         && \multicolumn{4}{c|}{ Models}  \\\cline{2-6}
         Defense& Apply &InceptionV3 & ResNet50 & MobileNet & VGG16 \\\hline
         JPEG \cite{guo2018countering}  & Only  & 9.82 &0.0&0.0 &0.0\\
         (ICLR '18) & Inference& &&&\\\hline
         JPEG DNN \cite{liu2019feature}&Only   & 13.12&0.35&0 & 18.38\\ 
         (CVPR '19) & Inference& &&&\\\hline\hline
          \textbf{*}BaRT k =5 \cite{BOT2019}[&Finetune$+$ & NA& 16.0&NA&NA\\ 
          (CVPR '19) & Inference& &&&\\\hline
         \textbf{*}BaRT k =10 \cite{BOT2019} &Finetune$+$&NA &36.0 &NA& NA \\
         (CVPR'19) & Inference& &&&\\\hline
          \our &Only  & \textbf{19.65}&\textbf{41.94}&\textbf{35.6}&\textbf{21.5} \\
          Our & Inference& &&&\\\hline\hline
    \end{tabular}
    \label{tab:imagenet_bpda}
\end{table}{}
\vspace{-0.5cm}
\vspace{-0.3cm}

To further ensure the effectiveness of proposed defense strategy, we also evaluate it against EOT (Expectation over Transformation), where attacker aims to capture the randomness in the transform by performing the transformation multiple times and use the average gradient. However, owing to presence of randomness at different steps of the transformation, the expectation over transformations fails to capture the randomness in the transform, even with as large as 100 runs. The InceptionV3 model achieved an accuracy of $40.19 \%$ 
over 100 steps of EoT. Further, we also investigated the combination of the two i.e.  BPDA +EoT, where the defense achieved an accuracy of $19.1 \%$ similar to BPDA alone.
\\

\noindent \textbf{Effect on Clean Sample Accuracy.} In the absence of \emph{detection} of adversarial examples, 
the input transformation strategies are applied to both clean and adversarial samples, adversely effecting the performance on clean samples. To cope with this drop, network fine-tuning is done with the proposed defense, prior to applying defense at the inference time. We report that with GraCIAS, performance on clean samples drop by ~16 and ~23 $\%$ for Inception and ResNet50 models on ImageNet without fine-tuning. However, this drop can be reduced with network fintuning as suggested by BaRT that achieves $65\%$ on clean samples against original clean sample accuracy of $76\%$,  \emph{i.e.} $11\%$ drop. Due to limited hardware resources, we verified this on ResNet model for CIFAR10, where the fine-tuned model regained the drop by $8\%$, suggesting similar benefits for ImageNet dataset on finetuning with our defense strategy.
\section{Ablation Study}
\vspace{-0.3cm}
We now investigate the choice of parameters in our defense strategy. While the range for percentage of variance in data to be retained as well as range of  number of filters is fixed across all the experiments, the parameter like filter size can depend on the dimension of the image. Also, performance of image operations to create the corrupted image set is evaluated under adaptive attack setting to validate the effectiveness of random filters over others.

\noindent \textbf{Effect of Filter Size.}
The goal of our filtering operation is to develop a diverse yet informative set of images to estimate a subspace that retains task-relevant information. This  effect can be verified from the results in Table \ref{tab: ablation} ImageNet datasets for two different perturbation magnitudes across three different filter sizes.
\begin{table}[h]
\centering
 \caption{\textbf{(Left)} Effect of selecting different transforms to create the set of corrupted image given in Eq. \ref{eq:ptrb_images} needed for our $\our$ defense. \textbf{(Right)} Effect of filter size on defense performance on ImageNet dataset at different perturbation levels under adaptive adversary (BPDA+PGD) with 100 iterations}
{%
  \begin{tabular}{|c|c|}
     \hline
      Operation & Defense Accuracy \\\hline
         Gaussian Filter &17.11  \\
         Affine Transformation &  11.86\\
         Symmetric Transformation &  7.29\\
         \hline
     \end{tabular}}%
\qquad
{%
    \begin{tabular}{|c|p{1.2cm}|p{1.2cm} |}
     \hline
      Filter Size & $\epsilon = 8$ & $\epsilon =16$\\\hline
      3 &17.11 & 16.41\\
      5 & 17.43&18.90 \\
      7 &22.38&19.65 \\\hline
    \end{tabular}
     }
     \label{tab: ablation}
\end{table}%

The results are reported in the adaptive attack setting (PGD +BPDA) to show the effect in stronger attack scenario. The results also point to the fact that in most real world applications, higher image resolutions are encountered, the choice of filter choice is not difficult.
\\
\begin{wraptable}[15]{r}{0.7\textwidth}
  \centering
    \caption{ImageNet-10K: Performance of simple defenses with GraCIAS used as pre-processing at the inference time on various models with $\epsilon =16$ under PGD10 attack. The boost in the performance is indicative of GraCIAS ability to restore the image details making it easier for much simpler defense like JPEG and BitDepth to defend the attack}
   
    \begin{tabular}{cccc}
    \hline
    Defense & \multicolumn{3}{c}{Accuracy}\\ & InceptionV3 \; & ResNet50 \; & VGG16 \\\hline
    JPEG &5.58 & 39.2 & 18.9 \\
    GraCIAS $\rightarrow$ JPEG & \textbf{44.26}&\textbf{45.14} & \textbf{31.79}\\
    BitDepth & 0.42&39.38 &27.95 \\
    GraCIAS $\rightarrow$ BitDepth &\textbf{25.97}&\textbf{43.79}&\textbf{29.89}\\
       \hline 
    \end{tabular}
    \label{tab:add_m2m}
\end{wraptable}
\noindent\textbf{Random Filters vs. Other Transforms.} We evaluated the effect of different image transforms to create the set of images in Eq. \ref{eq:ptrb_images} required for defining the subspace. The results for the same are given in the Table \ref{tab: ablation}. The performance suffers a significant drop with affine and symmetric transformation, due to the likely reason that the typically high frequency adversarial noise are still retained after such transformations. 

\noindent \textbf{GraCIAS as Pre-processing Prior to Other Defenses.}
JPEG compression and Bit depth reduction are simple input transformations that are very easy to incorporate in real world systems as they are model agnostic and can be implemented in hardware as well, however they become ineffective in presence of large perturbation as well as adaptive adversary. Since the proposed defense is also inexpensive in terms of computation resources for its implementation, it can be used to complement these defenses to retain their defense capabilities. The results in Table \ref{tab:add_m2m} show improvements on ImageNet dataset under PGD attack.
\vspace{-0.25cm}

\section{Conclusion}
In this work, we proposed a simple randomized linear subspace-based input defense approach that is applied at inference time to mitigate the effect of adversarial noise. The proposed approach achieved state-of-the-art results on ImageNet dataset across four different deep classification networks. The proposed defense is extensively evaluated on attacks with different strength and magnitude and is shown to be effective in all cases.
\bibliographystyle{splncs04}
\bibliography{Main_arxiv}

\end{document}